# Feature Selection Approach with Missing Values Conducted for Statistical Learning - A Case Study of Entrepreneurship Survival Dataset

Diego Nascimento
dcn@usp.br

Anderson Ara
anderson.ara@icmc.usp.br

Francisco Louzada
louzada@icmc.usp.br

**ABSTRACT**

In this article, we investigate the features which enhanced discriminate the survival in the micro and small business (MSE) using the approach of data mining with feature selection. According to the complexity of the data set, we proposed a comparison of three data imputation methods such as mean imputation (MI), k-nearest neighbor (KNN) and expectation maximization (EM) using mutually the selection of variables technique, whereby t-test, then through the data mining process using logistic regression classification methods, naïve Bayes algorithm, linear discriminant analysis and support vector machine hence comparing their respective performances. The experimental results will be spread in developing a model to predict the MSE survival, providing a better understanding in the topic once it is a significant part of the Brazilian' GPA and macro economy.

***Keywords—***feature selection, missing values, classification, entrepreneurship survival

## 1. INTRODUCTION

The increased amount of data in multidisciplinary, daily collections, demands new statistical strategies to manage and evaluate the hypothesis in a timely manner. Strategies such as dimensionality reduction, data mining and visualization are of importance when dealing with high-dimensional data and unstructured data. For this, a new field in statistics is rising called statistical learning which covers that as toolbox that aims to explore datasets, exploring the relationships among the structures from this data, considering it labeled or not. (FRIEDMAN, HASTIE & TIBSHIRANI, 2001).

Even in the information era it's common to find faulted database, presenting missing values, in this case one needs some kind of treatment in order to obtain some modelling out of it, a few prediction models are able to achieve it purpose. Regression or classification models with an incomplete dataset are very likely to use an imputation stage meanwhile to routine the feature selection step.

Imputation technics has been studied since the 30's, in despite of, Little and Rubin (2002) emphasize that the evolution of the imputation methods occurs middle 80's with the Bayesian approach adapted to this type of problems and using the computational facilities to simulate posterior distributions in order to have received extensive development, techniques such as expectation maximization is becoming nowadays a standard tools showing great results in the literature.

In the modeling' context, a feature selection preprocessing was adopted remains a great importance, in the performance of the classification models context, aiming reducing the dimension of the original feature space by determining a reduced subset of the most relevant features for a given problem, and making the classifier more efficient. The feature selection adopted method was t-Test, which utilizes data distribution as a key property for selecting variables identifying the actually influent features and better interpreting the classification models built as, for example, Logistic Regression, naïve Bayes, Linear discriminant analysis, and support vector machine (Fernández-Delgado, 2014).

These processes compose of pre-processing, feature selection, classification, and evaluation. As an application of the theme, was adopted a very explored in the literature topic such as enterprise survival of micro and small businesses, and the impact of the pro-entrepreneurship policies, though the characterization of their respective measures as discriminant factors still presented as underexplored topic in the Brazilian scene.

With motivation of the problem presented by Iglesias (2009), the author through a list obtained at the SEBRAE / RN and the State Board of Trade of Rio Grande do Norte (JUCERN), the enterprises registered from 2000 to 2002 were investigated through a survey. According to the theoretical framework of entrepreneurship survival, defined in three groups of factors - Structure Corporate Finance, Preparation for Entrepreneurship and Enterprising form - the factors were studied to determine whether they are discriminating or not for survival in the companies presented in Natal/RN.

This paper, therefore, aims analysis the most significant factors allowing the assessment of the pro-entrepreneurial policies and found that these policies have proved ineffective to prevent mortality companies. In addition, and most of all, to indicate the most significant factors found it was suggested change in these policies through a perspective of the application of binary classification techniques.

The paper is organized as follows. In section 2, we present the recurrent methodologies in the literature about the adopted missing data and imputation methods, classification techniques, and feature selection. Subsequently, section 3, we expose some applications, using an artificial and real-world dataset, and comparing the classifiers adopting through the 5-fold cross validation in order to represent those statistical results as efficiency and consistency. Finally, in section 4, we discuss the contribution of the work, and eventually, we conclude about the importance and challenges of the topic under study today.

## 2. METHODS

### 2.1. MISSING DATA AND IMPUTATION METHODS

The database on the survival of companies in Natal / RN present many zeros, and for lack of information, it is not possible to discern when the response to the characteristic actually assigns the zero score differentiating from the missing information where the researcher also assigned the note zero.

In order for the subsequent statistical inference to be valid, it is essential to inject the correct degree of randomness into the imputations and to incorporate this uncertainty in calculating standard errors and confidence intervals for the parameters of interest (LITTLE & RUBIN, 2002). Therefore, It was decides to deal equally to the zero grades as missing data, and in other to solve this this paper used three imputation techniques Mean Imputation (MI), k-nearest neighbor (KNN) and expectation maximization (EM).

    I.    Mean Imputation (MI): Mean Imputation makes only a trivial change in the correlation coefficient and no change in the regression coefficient. It adds no new information to the data but only increases the sample size. The objective of it reflects on the effect of

increasing the sample size is to increase the denominator for computing the standard error, thus reducing the standard error.

II. K-nearest neighbor (KNN): This method the missing values of a variable are imputed considering a given number of variables that are most similar to the instance of interest, using only local information by imputing information considering similar groups. The similarity of two instances is determined using a distance function (LITTLE & RUBIN, 2002).

III. Expectation maximization (EM): This method estimates the covariance matrix and impute values, preserving the relationship with other variables, which is vital if one goes on to use something like Factor Analysis or Regression. Still underestimate the standard error, however, so once again, this approach is only reasonable if the percentage of missing data are very small. It is an interactive procedure in which it uses other variables to impute a value (Expectation), then checks whether that is the value most likely (Maximization) (LITTLE & RUBIN, 2002).

In general, the Mean Imputation (MI) is considered one of the simplest and widely imputation technique presented in the literature, and if compared with, K-nearest Neighbor (KNN), which presents a slightly greater complexity, where it tries to group the data and impute values by it similar. The expectation maximization (EM) technique brings even a greater complexity, through simulation, estimates the parameters through probabilistic models for imputation of the incomplete data, having a higher computational cost.

## 2.2. CLASSIFICATION METHODS

The classification' data analysis task is to adaptive models or classifiers to construct or to predict categorical labels. The goal of classification is to accurately predict the target class for each case in the data. The methods used for classification first predict the probability of each of the categories of a qualitative variable, as the basis for making the classification (FRIEDMAN; HASTIE; TIBSHIRANI, 2001).

I. Logistic Regression (RL): In statistics, logistic regression is a regression model where the dependent variable is categorical. This article covers the case of binary dependent variables—that is, where it can take only two values.

II. Naïve Bayes (NB): The classifier is based on a maximum likelihood training that can be applied through an evaluating a closed-form expression, not adopting an iterative approach which would have a high cost operation, which its performance is based on a family of simple probabilistic classifiers through the application of Bayes' theorem considering independence between the characteristics.

III. Linear Discriminant Analysis (LDA): Introduced by Fisher (1936), the discriminant analysis is based on the construction of one or more linear functions involving the explanatory variables. This technique has the following assumptions: (1) the covariance matrices of each classification subset are equal. (2) Each classification group follows a multivariate normal distribution.

IV. Support Vector Machine (SVM): The associated learning algorithms that analyze the data, used for classification a dynamic of segregation, and have it represented as points in space, mapped so the examples of the separate categories are divided by a clear gap that is as wide as possible. Each tagged is marked as belonging or not to the category, this way the SVM training algorithm constructs a model that assigns new examples to the category, becoming a non-probabilistic binary linear classifier (VAPNIK, 1998).

## 2.2. FEATURE SELECTION

The Student's t-test was used to select the variables which demonstrated statistical significance at the characteristics, considering a level of significance as 0.05, whose maximum mean value in the difference between groups and as well presenting minimum variability.

## 3. RESULTS

In this section we present the results. In section 3.1 we describe the real dataset and section 3.2 presents the application of the classification and imputation methods.

### 3.1. Dataset Detail

| Table 1 – Factors affecting the survival of MSE ||
|---|---|
| **Topics** | **Description** |
| Personal features | The ability of human interaction, communications skills and knowledge technics. |
| | Managerial and leadership skills. |
| | Self employment, previous work experience and entrepreneurial skills (organization, motivation and communication skills). |
| | Management competence. |
| | Entrepreneurs skills (financial management and personnel management). |
| | Management styles. |
| | Educational background and technical training. |
| Business features | Enterprise operation (lowering cost of production and distribution). |
| | Enterprise location. |
| | Economic factors. |
| | Risk and returns analysis. |
| | Technical performance, cost, and alternative technologies. |
| | The local environment. |
| | Business plan analysis. |
| | Competitive strategy (fields as marketing, management, human resources and finance). |
| | Strategic Networks analysis. |
| | Specialization in products and services. |
| | Innovation. |
| | Adopted strategy as such as target market and the internal business controls. |
| Financial Support | Credit access analysis. |
| Institutional Support | Business assistance. |

| Personal and Business features | Staff motivation in the business environment. |
|---|---|
| Personal and Business features | Creative business solution (Strategies). |
| Personal and Business features | Hierarchy, bureaucracy and complacency, greed, agility and innovation. |
| Personal and Business features | Interpersonal network, financial resources, personal skills and social resources. |
| Features Business and Personal, and Financial Support | Lack of capital or administrative capacity. |
| Personal Features and Financial Support | Professional training and has previous work experience. |
| Personal Features and Financial Support | Customer demographics, previous experience in business and access to capital. |
| Personal Features, Financial Support and Institutional Support | Joint efforts of universities and SMEs, support through Research and Development and the creation of networks of large and diversified business activities. |
| Features Business and Personal, and Financial Support | Entrepreneur talent (personal characteristic), technology and capital (physical, tangible) and know-how (knowledge and experience). |
| Features Business and Personal, Financial Support and Institutional Support | Financial support and good management techniques, which creates a favorable economic environment, preparing of the entrepreneur with hand-intensive training and good facilities. |

## 3.2. MICRO AND SMALL BUSINESS SURVIVAL DATASET

The database came from Iglésias dissertation research, which the total number of enter was 981, thus a sample of 149 companies responded, in which 52 (or 35%) had deactivated while 97 (or 65%) had activated status in the Commercial Board of the State Rio Grande do Norte (JUCERN). The grades attributed as zeros were removed, because of their inconsistencies, and may be related to the lack of information of the respondents, and then in this search it was considered simply as missing data.

Analyzing initially some results, the question that obtained the highest average is related to the business' future project planification (Q37), in contrast the lowest average is the one related to an existing team with specific training in management, assisting the company' decisions (Q25). Thus further, the only question which presented no missing data was upon the accomplishment of deadline guided by director (Q44), meanwhile the question regarding the granting of some type of financing made during the business' implementation (Q16) presented the biggest missing numbers.

In the general overview, the heatmap in Figure 1 graphically shows the dispersion of the missing values and its related questions, in which the brighter color represents the existence of the missing value and the darker color represents the existence of a missing, thus it observed value. It is worth mentioning that the axis of the ordinates represents the companies and the axis of the abscissa represents the attributes (question), and is in descending order by the number of missing values.

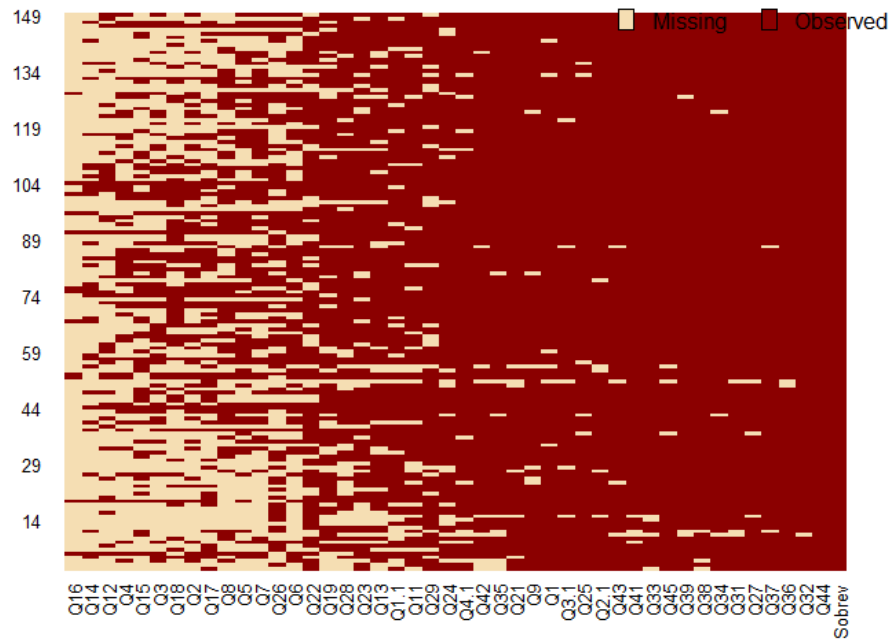

Figure 1. Heatmap of the missing values

### 3.3 GENERAL RESULTS

Three imputation methods was adopted respectively; Mean Imputation (MI), k-nearest neighbor (KNN), with *k=3*, combined with all four classifiers; Logistic Regression, naïve Bayes, Discriminant Analysis, and Support Vector Machine. In order to the number of missing data, different values of *k* do not converge in KNN imputation method. For statistical inference propose, the 5-fold cross validation technique was also implemented, and the performance of each classifiers was measured by the mean of the folds considering it accuracy (ACC), area under the ROC (AUC) curve, sensitivity (SEN) and specificity (SPE).

For generalization of modeling results, the mean and standard deviation of the 5-fold cross validation method were calculated, for each performance indicator, and compared the techniques as shown in the Table 1.

The Naïve Bayes classifier, combined with the KNN imputation method, obtained higher prosecution considering all measures of classifier performance, represented by higher mean, ought was not the best one. However, Support Vector Machines performs with good predictive performance as well.

| Methods | | ACC | | AUC | | SEN | | SPE | |
|---|---|---|---|---|---|---|---|---|---|
| | | Mean | SD | Mean | SD | Mean | SD | Mean | SD |
| Logistic Regression | MI | 0.6216 | 0.0758 | 0.678 | 0.1044 | 0.4549 | 0.0593 | 0.7321 | 0.1089 |
| | KNN | 0.6388 | 0.0293 | 0.7165 | 0.0351 | 0.4815 | 0.1107 | 0.7506 | 0.1152 |
|  | EM | 0.3494 | 0.0793 | 0.7151 | 0.0845 | 0.3494 | 0.0793 | NaN | NaN |
| Linear Discriminant Analysis | MI | 0.6385 | 0.0695 | 0.7004 | 0.1063 | 0.4912 | 0.1008 | 0.7380 | 0.1035 |
|  | KNN | 0.6552 | 0.0356 | 0.7196 | 0.0518 | 0.523 | 0.1462 | 0.7713 | 0.1157 |
|  | EM | 0.8053 | 0.0551 | 0.8899 | 0.0381 | 0.7390 | 0.1654 | 0.8654 | 0.1053 |
| Support Vector Machine | MI | 0.6308 | 0.0852 | 0.7676 | 0.0924 | 0.6189 | 0.3518 | 0.688 | 0.1734 |
|  | KNN | 0.6786 | 0.1288 | 0.786 | 0.0893 | 0.6198 | 0.3306 | 0.7322 | 0.1717 |
|  | EM | 0.7653 | 0.0995 | 0.8627 | 0.041 | 0.69 | 0.2945 | 0.7727 | 0.1632 |
| Naive Bayes | MI | 0.723 | 0.1308 | 0.7917 | 0.1032 | 0.5540 | 0.2272 | 0.8158 | 0.1192 |
|  | KNN | 0.7566 | 0.0617 | 0.8089 | 0.0704 | 0.6138 | 0.148 | 0.8486 | 0.1006 |
|  | EM | 0.7317 | 0.0966 | 0.8212 | 0.087 | 0.6242 | 0.1885 | 0.8019 | 0.0911 |

The LDA classifier, combined with the EM imputation method, obtained higher prosecution considering all measures of classifier performance, represented by higher mean. The performance of these methods should be verified graphically through the ROC curves, they assist in the process of performance analysis of the classification techniques for each imputation method as presented Figure 2.

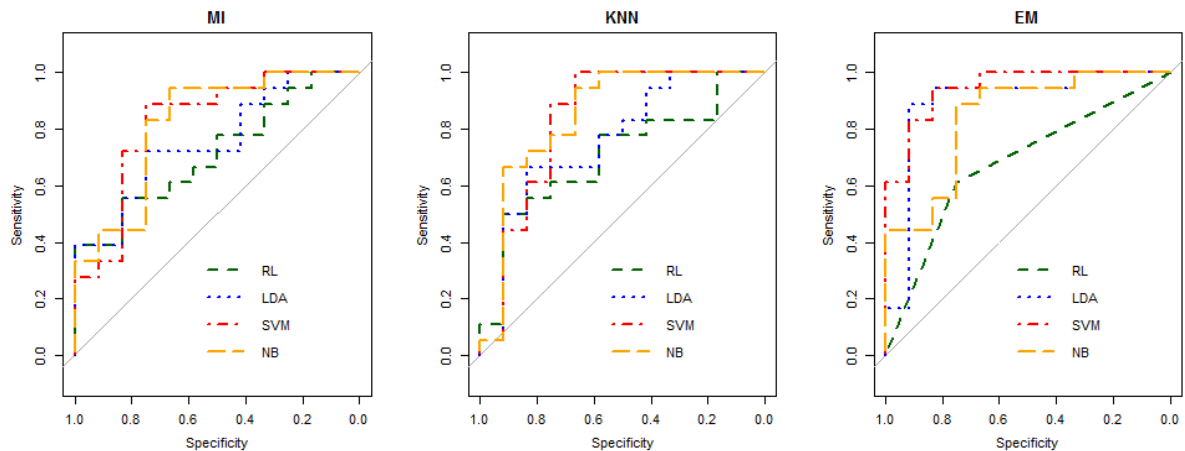

Figure 2. ROC curve and AUC area of the three imputation methods.

## 4. FINAL COMMENTS

The results presented in this paper show that the nonlinear classifiers had an outperformance comparing to that linear, as elements presents that Support Vector Machine and Naïve Bayes act better in both datasets, the real and artificial one. Considering that nonlinear classifiers have more flexibility and may present a higher performance, the results show that Naïve Bayes classifier had the best result for the entrepreneurship application.

In this paper we combined and compared the Mean Imputation, KNN and Expectation Maximization techniques as an alternative to implement the classifiers' algorithms as a full data set, and also to improve their performances. An interesting point to be noted is that by comparing the parameters used in the KNN imputation technique, in the linear decision surfaces, the KNN grouping the 3 closest elements (KNN3) performed better for the logistic regression and linear discriminate analysis. In the other hand, for the nonlinear the KNN using only the first closest element (KNN1) presented a better performance, ought the best performance was the combination with the Expectation Maximization techniques and Linear Discriminant Analysis.

For future works, we are considering to investigate further explanations about the different performances by it parameterization, and also to be included simulations part in order to explain the results obtained and others applications for features selections.